\documentclass{article}
\usepackage{spconf,amsmath,graphicx,hyperref}
\setkeys{Gin}{draft}
\usepackage{booktabs}
\usepackage{subcaption}

\title{CMAMBADEPTH: SELF-SUPERVISED MONOCULAR DEPTH ESTIMATION
WITH CHANNEL MAMBA AND HYBRID ATTENTION}
\name{
Xuezhi Xiang\textsuperscript{\textup{1,2}}\sthanks{Corresponding author: xiangxuezhi@hrbeu.edu.cn\newline This work was supported in part by National Natural Science Foundation of China under Grant 62671193 and 62271160, in part by Heilongjiang Provincial Key R\&D Program Project under Grant 2026ZX01A14, in part by the Natural Science Foundation of Heilongjiang Provincial of China under Grant XQ2026F018, in part by the Fundamental Research Funds for the Central Universities of China under Grant 3072026LJ0802.},
Jiayao Liu\textsuperscript{\textup{1}},
Heqi Xiang\textsuperscript{\textup{3}},
Yuqi Hu\textsuperscript{\textup{1}},
Yiming Chen\textsuperscript{\textup{1}},
Shanjun Zhang\textsuperscript{\textup{4}}
}
\address{\parbox{\textwidth}{\centering
\fontsize{12}{14}\selectfont
\makebox[\linewidth][c]{\textsuperscript{1}School of Information and Communication Engineering, Harbin Engineering University, Harbin 150001, China}\\
\makebox[\linewidth][c]{\textsuperscript{2}Key Laboratory of Advanced Marine Communication and Information Technology, Harbin 150001, China}\\
\textsuperscript{3}Department of Computer Science, University of Toronto, Toronto ON M5S 2E4, Canada\\
\textsuperscript{4}Department of Computer Science, Kanagawa University, Kanagawa 221-8686, Japan
}}
\begin{document}
%
\maketitle
\begin{abstract}
Accurate monocular depth estimation serves as a core enabler for single camera scene understanding. However, existing self-supervised monocular depth estimation methods generally suffer from the bottleneck of inefficient cross-scale information interaction and difficulty in balancing local and global spatial modeling. In this paper, we propose CMambaDepth, a self-supervised framework that achieves efficient multi-scale feature fusion and fine-grained contextual modeling via channel-wise selective state propagation. Specifically, Bidirectional Channel Mamba (Bi-CMamba) aligns encoder features across scales and enables bidirectional information exchange among ordered scale groups. Unidirectional Channel Mamba (Uni-CMamba) progressively aggregates decoder features and retains fine-grained scale groups through a group selection mechanism for subsequent fusion. Furthermore, a Hybrid Attention Module (HAM) is introduced to combine large-kernel local context and Manhattan self-attention for complementary spatial modeling. Experimental results demonstrate that our method achieves highly competitive performance. Specifically, our model achieves an AbsRel of 0.094 and an RMSE of 4.156 on KITTI, and an AbsRel of 0.140 on DDAD. In the zero-shot cross-dataset generalization test on NYUv2, it attains an AbsRel of 0.232, outperforming the baseline RA-Depth by 7.2\%.
\end{abstract}
\section{Introduction}
\label{sec:intro}

\noindent Monocular depth estimation underpins navigation and robotic perception. Supervised multi-scale learning \cite{ref1} lays the foundation for label-free stereo reconstruction \cite{ref2} and left-right consistency constraints \cite{ref3} without depth labels. Monocular video training \cite{ref4}, minimum reprojection, and auto-masking \cite{ref5} extend the scope of self-supervised learning. SC-DepthV3 \cite{ref6} addresses dynamic scenes, while DCPI-Depth \cite{ref7} incorporates dense correspondence priors.

Feature aggregation plays a central role in preserving geometric structures for monocular depth estimation. MonoViT \cite{ref8} models global context with vision transformers, and RA-Depth \cite{ref9} learns resolution-adaptive feature representations. TSUDepth \cite{ref11} explicitly models temporal uncertainty. However, most existing approaches still suffer from inefficient cross-scale feature fusion, lacking a selective mechanism to adaptively integrate complementary cues while mitigating inherent scale ambiguity. This gap therefore motivates the exploration of selective cross-scale interactions.

Mamba \cite{ref14} makes state space parameters input-dependent. VMamba \cite{ref15} and MambaDepth \cite{ref16} extend this mechanism to vision and depth, respectively. Yet existing visual Mamba models fail to jointly capture fine-grained local details and model global context within a single compact block. To address this issue, we design HAM that combines large kernel attention (LKA) \cite{ref17} and Manhattan self-attention (MaSA) \cite{ref18} on the aligned scale-specific decoder features.

\begin{figure*}[t]
  \centering
\includegraphics[
  draft=false,
  width=0.80\textwidth
]{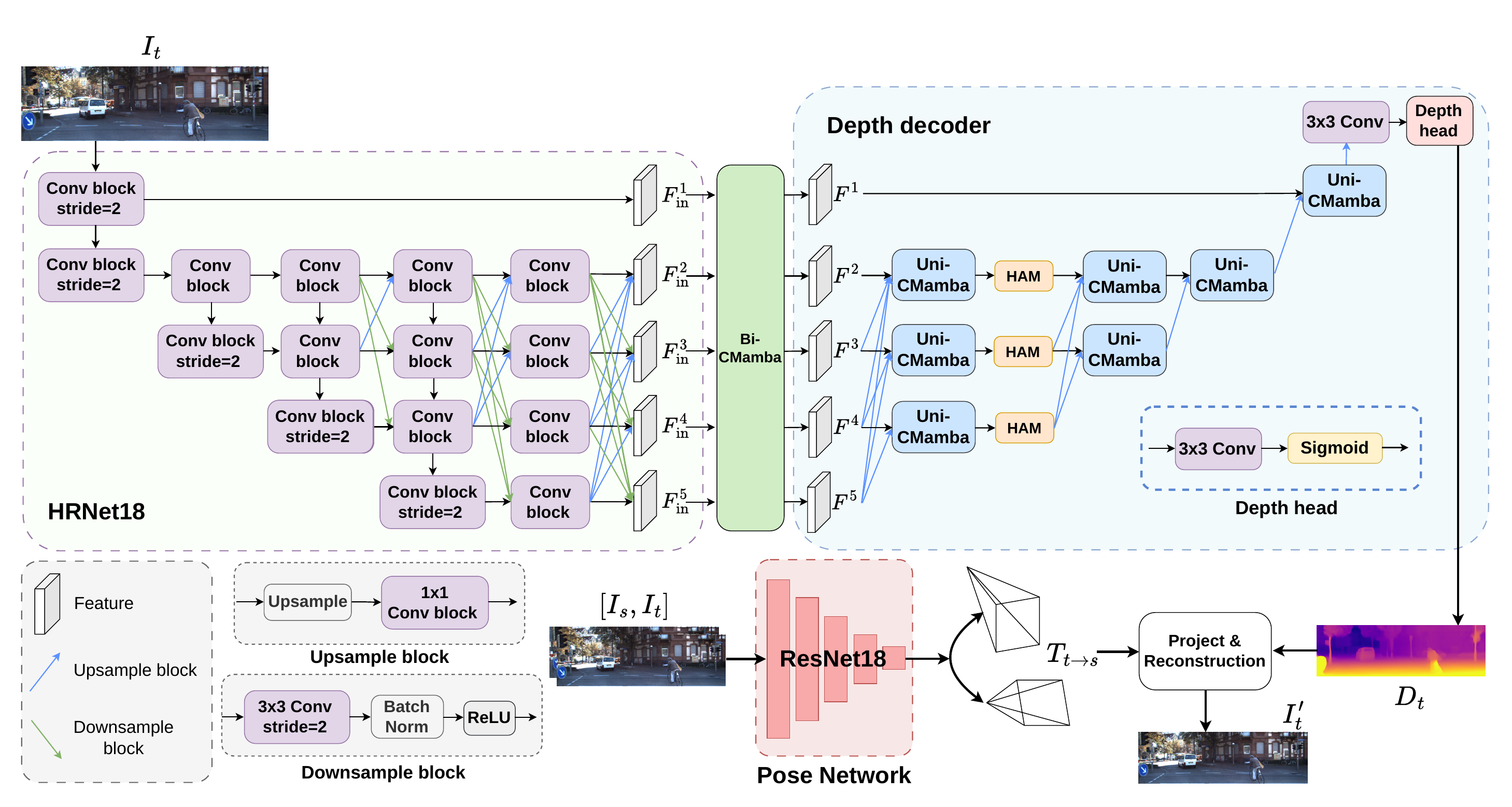}
  \caption{Overall architecture of the self-supervised depth estimation framework. Bi-CMamba connects the HRNet18 encoder to progressive Uni-CMamba decoding. Three HAM instances enrich parallel streams after the first aggregation round.}
  \label{fig:large_framework}
\end{figure*}

We adopt RA‑Depth\cite{ref9} as the baseline, and our contributions are as follows:
\begingroup
\settowidth{\leftmargini}{\labelitemi}
\addtolength{\leftmargini}{\labelsep}
\begin{itemize}
\item We propose Bi-CMamba and Uni-CMamba, that recast multi-scale feature fusion as selective state propagation across ordered scale groups. Bi-CMamba enables bidirectional cross-scale information flow at the encoder–decoder bridge, while Uni-CMamba performs a single coarse-to-fine scan of decoder feature groups and preserves fine-grained scale groups.
\item We design HAM, which fuses a decomposed LKA
local branch and an axial MSA global
branch through learnable and broadcast channel
weights.
\item CMambaDepth attains AbsRel 0.094 on KITTI and 0.140 on DDAD, and reduces AbsRel by 7.2\% over RA-Depth in the zero-shot cross-dataset generalization test on NYUv2. 
\end{itemize}
\endgroup

\section{METHOD}
\label{sec:format}
\subsection{Overall architecture}
\label{ssec:baseline}

\noindent \hyperref[fig:large_framework]{Fig.\ \ref*{fig:large_framework}} shows the overall architecture of CMambaDepth. Built on RA-Depth \cite{ref9}, the network adopts a five-scale HRNet18 encoder \cite{ref19} and a ResNet18 pose encoder \cite{ref20} used only during training. Bi-CMamba exchanges information bidirectionally across encoder scales. Four Uni-CMamba stages fuse decoder features in a coarse-to-fine order, with three HAM instances deployed in parallel between the first two stages. Finally, a depth head produces the map from the fused features.

The depth and pose networks take monocular triplets $\{I_{t-1},I_t,I_{t+1}\}$ as input and output the depth map $D_t$ and the relative pose $T_{t\to s}$, respectively. For $s\in\{t-1,t+1\}$, the target view is reconstructed as
\begin{equation}
I'_t=I_s\langle\operatorname{Proj}(D_t,T_{t\to s},K)\rangle,
\label{eq:method_reconstruction}
\end{equation}

\noindent where $K$ denotes the camera intrinsic matrix, $\mathrm{Proj}(\,\cdot\,)$ projects
target pixels to the source view, and $\langle\cdot\rangle$ denotes
bilinear sampling.

\subsection{Bidirectional-channel Mamba}
\label{ssec:bidirectional}

\begin{figure}[t]
 \includegraphics[
  draft=false,
  width=\columnwidth
]{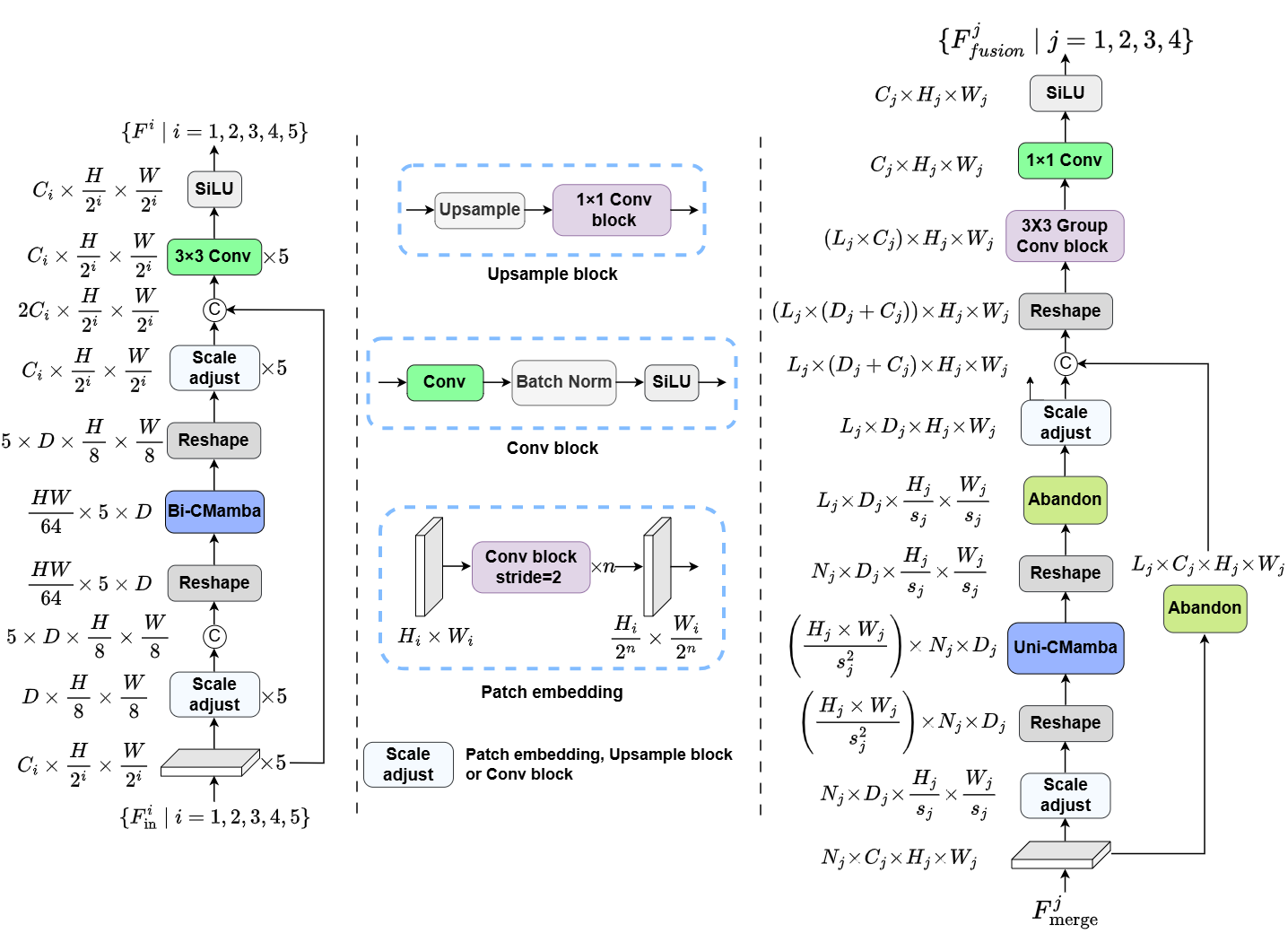}
  \caption{Architectures of Bi-CMamba (left) and Uni-CMamba (right).}
  \label{fig:comparison}
\end{figure}

\noindent As shown in \hyperref[fig:comparison]{Fig.\ \ref*{fig:comparison}} (left), the five-scale encoder features are first aligned to a common resolution of $(H/8)\times(W/8)$ and partitioned into groups, each containing $D$ channels. Inside a gated residual Visual State-Space (VSS) block \cite{ref15}, Bi-Mamba scans each group bidirectionally using input-dependent state-space parameters \cite{ref14}. The resulting outputs $F^i_{mamba}$ are upsampled back to the original resolution of the $i$-th scale and fused with the input features $F^i_{in}$ at that scale:
\begin{equation}
F^i=\operatorname{SiLU}\!\left(\operatorname{Conv}_{3\times3}\!\left(\operatorname{Concat}(F^i_{mamba},F^i_{in})\right)\right).
\label{eq:method_encoder_fusion}
\end{equation}

\subsection{Unidirectional-channel Mamba}
\label{ssec:unidirectional}

\noindent In \hyperref[fig:comparison]{Fig.\ \ref*{fig:comparison}} (right), the aligned decoder features $F^j_{merge}$ are first downsampled and then scanned once along the coarse-to-fine direction, yielding $F^j_{mamba}$. Spatial reshaping and group selection are applied before upsampling:
\begin{equation}
F^j_{out}=\operatorname{Upsample}\!\left(\operatorname{Abandon}\!\left(\operatorname{Reshape}(F^j_{mamba})\right)\right),
\label{eq:method_upsampling}
\end{equation}

\noindent where $\mathrm{Reshape}(\,\cdot\,)$ rearranges the spatial dimensions. Across the four successive rounds, the numbers of input groups are 5, 4, 3, and 2, and $\mathrm{Abandon}(\,\cdot\,)$ retains the last 4, 3, 2, and 1 groups, respectively, after which $\mathrm{Upsample}(\,\cdot\,)$ restores the features to the target resolution to obtain $F^j_{out}$. A grouped $3\times3$ convolution then performs intra-group mixing on $F^j_{out}$, and its output is concatenated with the retained input streams to form $F^j_o$, which is further mixed across groups by a $1\times1$ convolution followed by $\mathrm{SiLU}(\,\cdot\,)$:
\begin{equation}
F^j_{fusion}=\operatorname{SiLU}\!\left(\operatorname{Conv}_{1\times1}(F^j_o)\right).
\label{eq:method_decoder_fusion}
\end{equation}

\subsection{Hybrid attention module and training}
\label{ssec:hybrid_training}

\begin{figure}[t]
 \centering
 \includegraphics[
  draft=false,
  width=0.70\columnwidth
]{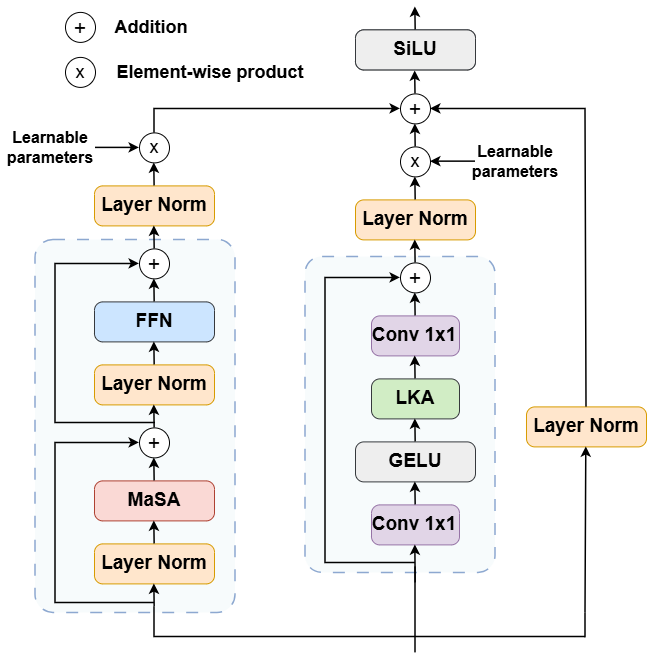}
  \caption{Hybrid Attention Module.}
  \label{fig:method_ham}
\end{figure}

\noindent As shown in \hyperref[fig:method_ham]{Fig.\ \ref*{fig:method_ham}}, HAM takes $F^i_{fusion}$ as input and consists of two residual branches, a global branch and a local branch. In the global branch, MaSA first applies axial self-attention along the horizontal and vertical directions to the LayerNorm-normalized features and adds the result through a residual connection. The output then passes through a residual feed-forward network (FFN) \cite{ref18} to produce $F^i_{global}$. The local branch produces $F^i_{local}$, computed as
\begin{equation}
\begin{aligned}
F^i_{local}={}&F^i_{fusion}+\operatorname{Conv}\!\left(\operatorname{LKA}\!\left(\operatorname{GELU}\!\left(\operatorname{Conv}(F^i_{fusion})\right)\right)\right).
\end{aligned}
\label{eq:method_local_branch}
\end{equation}

The two outer convolutions are both $1\times1$ convolutions. LKA \cite{ref17} decomposes the large $23\times23$ kernel into a $5\times5$ depth-wise convolution, a $7\times7$ depth-wise dilated convolution with a dilation rate of 3, and a $1\times1$ convolution. Weighted features are then added to the normalized original feature to obtain $X^i$. The final output is produced as:
\begin{equation}
X^i_{out}=\operatorname{SiLU}(X^i).
\label{eq:method_ham_output}
\end{equation}

Following RA-Depth \cite{ref9}, our minimum photometric reprojection loss combines an SSIM term with weight 0.85 and an L1 term with weight 0.15, and adopts the auto-masking strategy \cite{ref5}. The edge-aware smoothness loss regularizes the mean-normalized inverse depth, while the cross-scale depth consistency loss aligns corresponding regions across the low (L), medium (M), and high (H) resolutions. The overall loss is defined as:
\begin{equation}
\begin{aligned}
\mathcal{L}_{total}={}&\gamma\bigl(\mathcal{L}_p^L+\mathcal{L}_p^M+\mathcal{L}_p^H\bigr)
+\beta\bigl(\mathcal{L}_s^L+\mathcal{L}_s^M+\mathcal{L}_s^H\bigr)\\
&+\lambda\bigl(\mathcal{L}_{cs}^{LM}+\mathcal{L}_{cs}^{MH}\bigr),
\end{aligned}
\label{eq:method_training_loss}
\end{equation}

\noindent where $\mathcal{L}_p$, $\mathcal{L}_s$, and $\mathcal{L}_{cs}$ denote the photometric, smoothness, and cross-scale losses, respectively. Following the baseline setting, the weights are set to $\gamma=1$, $\beta=0.001$, and $\lambda=1$.

\section{EXPERIMENTS}

\label{sec:experiments}

\begin{table*}[t]
\centering
\caption{Comparisons on KITTI, DDAD, and NYUv2. \protect\textbf{Bold} and \protect\underline{underlined} values denote the best and second-best distinct results per dataset, including ties. Dashes (--) indicate unavailable or unverified values excluded from ranking.}
\label{tab:quantitative_comparison}
\small
\setlength{\tabcolsep}{2pt}
\renewcommand{\arraystretch}{1.08}
\begin{tabular*}{\textwidth}{@{\extracolsep{\fill}}ll*{8}{c}@{}}
\toprule
\textbf{Dataset} & \textbf{Method} & \shortstack{\textbf{AbsRel}$\downarrow$} & \shortstack{\textbf{SqRel}$\downarrow$} & \shortstack{\textbf{RMSE}$\downarrow$} & \shortstack{\textbf{RMSElog}$\downarrow$} & $\boldsymbol{\delta_1}\uparrow$ & $\boldsymbol{\delta_2}\uparrow$ & $\boldsymbol{\delta_3}\uparrow$ & \shortstack{\textbf{Params}\\(M)$\downarrow$} \\
\midrule
\raisebox{-27.828pt}[0pt][0pt]{\textbf{\shortstack[l]{KITTI\\$640\times192$}}} & MonoViT \cite{ref8} & 0.099 & 0.708 & 4.372 & 0.175 & 0.900 & 0.967 & 0.984 & 27.87 \\
 & MambaDepth \cite{ref16} & 0.097 & 0.706 & 4.370 & 0.172 & \textbf{0.907} & \textbf{0.970} & \textbf{0.986} & 30 \\
 & TSUDepth \cite{ref11} & \underline{0.095} & \textbf{0.605} & \underline{4.192} & \textbf{0.166} & \underline{0.905} & \textbf{0.970} & \textbf{0.986} & 128.47 \\
 & RA-Depth \cite{ref9} & 0.096 & 0.632 & 4.216 & 0.171 & 0.903 & 0.968 & \underline{0.985} & \textbf{9.98} \\
 & CMambaDepth\allowbreak\ (Ours) & \textbf{0.094} & \underline{0.616} & \textbf{4.156} & \underline{0.169} & \underline{0.905} & \underline{0.969} & \underline{0.985} & \underline{10.89} \\
\midrule
\raisebox{-21.996pt}[0pt][0pt]{\textbf{\shortstack[l]{DDAD\\$640\times384$}}} & TransDSSL \cite{ref21} & 0.151 & 3.591 & 14.350 & 0.244 & -- & -- & -- & -- \\
 & IntrLessMonoDepth \cite{ref22} & 0.151 & 3.049 & \textbf{13.431} & 0.237 & 0.814 & 0.929 & 0.966 & \underline{10.2} \\
 & RA-Depth \cite{ref9} & \underline{0.141} & \underline{2.877} & 13.734 & \underline{0.231} & \underline{0.827} & \underline{0.935} & \underline{0.971} & \textbf{9.98} \\
 & CMambaDepth\allowbreak\ (Ours) & \textbf{0.140} & \textbf{2.761} & \underline{13.550} & \textbf{0.229} & \textbf{0.829} & \textbf{0.936} & \textbf{0.972} & 10.89 \\
\midrule
\raisebox{-21.996pt}[0pt][0pt]{\textbf{\shortstack[l]{NYUv2\\KITTI-trained}}} & DCPI-Depth \cite{ref7} & 0.299 & 0.431 & 1.060 & 0.339 & 0.511 & 0.810 & 0.937 & -- \\
 & MonoViT \cite{ref8} & \underline{0.250} & 0.351 & 0.930 & \underline{0.293} & 0.594 & \underline{0.870} & \underline{0.961} & 27.87 \\
 & RA-Depth \cite{ref9} & \underline{0.250} & \underline{0.286} & \underline{0.843} & \underline{0.293} & \underline{0.605} & 0.861 & 0.955 & \textbf{9.98} \\
 & CMambaDepth\allowbreak\ (Ours) & \textbf{0.232} & \textbf{0.240} & \textbf{0.782} & \textbf{0.277} & \textbf{0.625} & \textbf{0.877} & \textbf{0.962} & \underline{10.89} \\
\bottomrule
\end{tabular*}
\end{table*}

\subsection{Datasets and implementation}

\label{ssec:experiment_1}

\noindent On KITTI \cite{ref23}, our model is trained for 35 epochs with a batch size of 11 and an initial learning rate of \(1\times10^{-4}\). On DDAD \cite{ref24}, the model is trained for 60 epochs with a batch size of 4. The initial learning rates of the pose encoder, depth encoder, and depth decoder are set to \(1\times10^{-4}\), \(1\times10^{-5}\), and \(6\times10^{-5}\), respectively. For NYUv2 \cite{ref25}, the model trained on KITTI is directly evaluated in a zero-shot cross-dataset setting without any fine-tuning. Our model is implemented in PyTorch, and all experiments are conducted on a single RTX 4090 GPU. We adopt Adam \cite{ref26} as the optimizer and initialize both the depth and pose encoders with ImageNet-pretrained weights \cite{ref27}. The data augmentation strategy and remaining loss settings follow RA-Depth \cite{ref9}. We adopt four error metrics, namely AbsRel, SqRel, RMSE, and log-depth RMSE (RMSElog), together with three accuracy metrics $\delta_1$, $\delta_2$, and $\delta_3$ under the ratio thresholds 1.25, $1.25^2$, and $1.25^3$.

\begin{table*}[t]
\centering
\caption{KITTI component ablation with Bi-CMamba, Uni-CMamba, and HAM. Check marks indicate enabled modules. \protect\textbf{Bold} denotes the best value for each metric across all configurations, including ties.}
\label{tab:component_ablation}
\small
\setlength{\tabcolsep}{1.5pt}
\renewcommand{\arraystretch}{1.08}
\def\modulecheck{\raisebox{1pt}{\rotatebox{-45}{\rule{2.4pt}{0.45pt}}}\kern-1pt\raisebox{0.3pt}{\rotatebox{60}{\rule{6pt}{0.45pt}}}}
\begin{tabular*}{\textwidth}{@{\extracolsep{\fill}}l*{12}{c}@{}}
\toprule
\textbf{Method} & \shortstack{\textbf{Bi-}\\\textbf{CMamba}} & \shortstack{\textbf{Uni-}\\\textbf{CMamba}} & \textbf{HAM} & \textbf{AbsRel}$\downarrow$ & \textbf{SqRel}$\downarrow$ & \textbf{RMSE}$\downarrow$ & \textbf{RMSElog}$\downarrow$ & $\boldsymbol{\delta_1}\uparrow$ & $\boldsymbol{\delta_2}\uparrow$ & $\boldsymbol{\delta_3}\uparrow$ & \shortstack{\textbf{Params}\\\textbf{(M)}} & \textbf{GFLOPs} \\
\midrule
\textbf{Baseline} &  &  &  & 0.096 & 0.632 & 4.216 & 0.171 & 0.903 & 0.968 & \textbf{0.985} & \textbf{9.98} & \textbf{10.78} \\
 & \modulecheck &  &  & 0.096 & 0.624 & 4.169 & 0.170 & 0.903 & \textbf{0.969} & \textbf{0.985} & 10.31 & 12.26 \\
 &  & \modulecheck &  & 0.095 & 0.621 & 4.174 & 0.170 & 0.903 & 0.968 & \textbf{0.985} & 10.24 & 12.75 \\
 &  &  & \modulecheck & 0.096 & 0.624 & 4.171 & 0.170 & 0.904 & \textbf{0.969} & \textbf{0.985} & 10.30 & 11.10 \\
\midrule
 & \modulecheck & \modulecheck &  & 0.095 & 0.619 & 4.169 & \textbf{0.169} & 0.904 & \textbf{0.969} & \textbf{0.985} & 10.57 & 14.21 \\
 & \modulecheck &  & \modulecheck & 0.095 & 0.618 & 4.163 & \textbf{0.169} & \textbf{0.905} & 0.968 & \textbf{0.985} & 10.62 & 12.57 \\
 &  & \modulecheck & \modulecheck & 0.095 & 0.620 & 4.161 & \textbf{0.169} & \textbf{0.905} & 0.968 & \textbf{0.985} & 10.56 & 13.05 \\
 \textbf{Ours} & \modulecheck & \modulecheck & \modulecheck & \textbf{0.094} & \textbf{0.616} & \textbf{4.156} & \textbf{0.169} & \textbf{0.905} & \textbf{0.969} & \textbf{0.985} & 10.89 & 14.52 \\
\bottomrule
\end{tabular*}
\end{table*}

\begingroup\setlength{\intextsep}{0pt}
\begin{figure}[!ht]
\centering\captionsetup{skip=2pt}
\includegraphics[
  draft=false,
  width=1.00\columnwidth
]{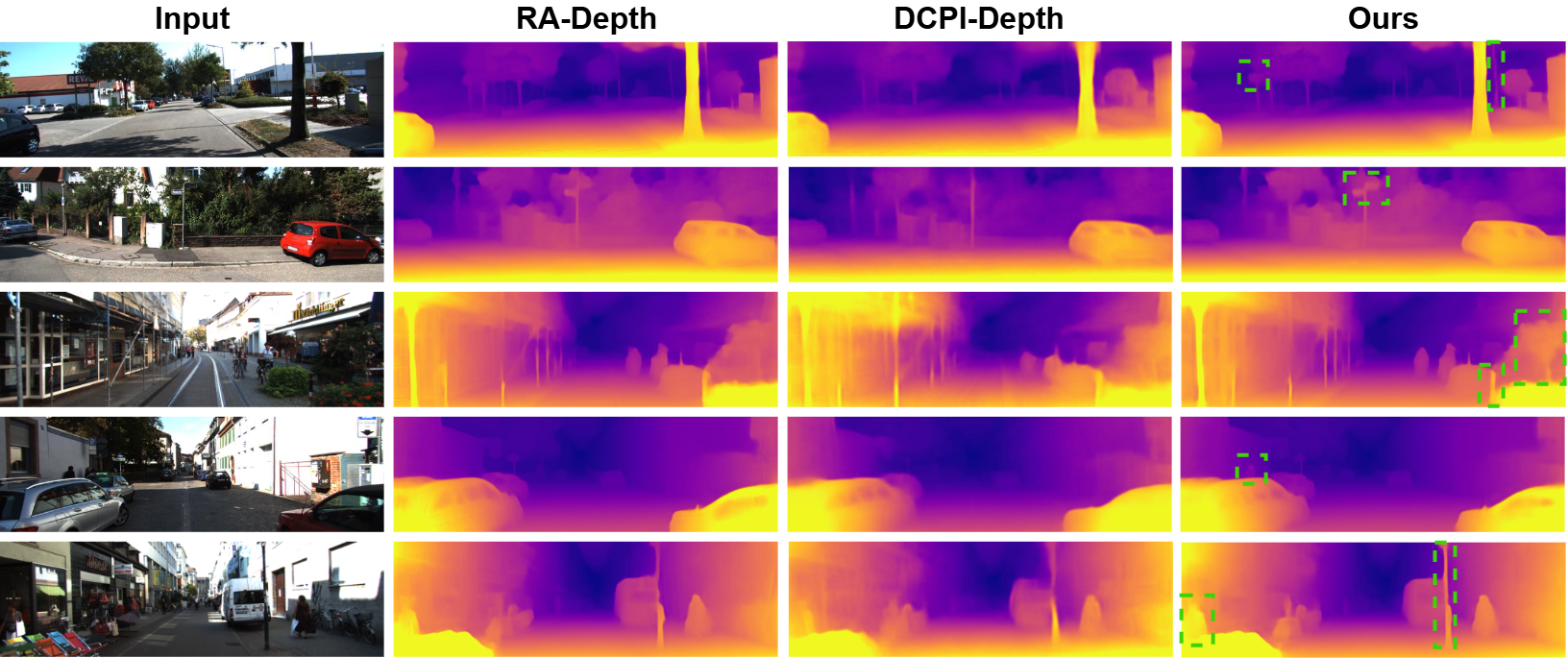}
\caption{KITTI qualitative comparison. Dashed boxes highlight signs, pedestrians, poles, and vegetation.}
\label{fig:experiments_4}
\end{figure}
\endgroup

\subsection{Quantitative comparisons}

\label{ssec:experiment_2}

\noindent \hyperref[tab:quantitative_comparison]{Table~\ref*{tab:quantitative_comparison}} provides Comparisons on KITTI, DDAD, and NYUv2. On KITTI, CMambaDepth\allowbreak\ (Ours) reduces the baseline AbsRel and RMSE by 2.1\% and 1.4\%, reaching 0.094 and 4.156, respectively. With only 10.89 M parameters, our model already achieves a lower RMSE than MonoViT \cite{ref8}.On DDAD, CMambaDepth\allowbreak\ (Ours) improves over RA-Depth on all seven metrics. Compared with TransDSSL (2022) \cite{ref21}, AbsRel and RMSE decrease from 0.151 and 14.350 to 0.140 and 13.550, respectively. IntrLessMonoDepth (2025) \cite{ref22} attains a slightly lower RMSE, whereas our method is superior on  AbsRel and SqRel. Overall, our method achieves competitive results on both datasets with a compact model, which we attribute to the efficient selective cross-scale fusion performed by Bi-CMamba and Uni-CMamba, and the low-cost contextual modeling provided by HAM.

Under the zero-shot setting without any fine-tuning on NYUv2, our method reduces AbsRel, SqRel, and RMSE over RA-Depth, achieving to relative reductions of 7.2\%, 16.1\%, and 7.2\%, respectively. Meanwhile, CMambaDepth (Ours) outperforms both MonoViT (2022) \cite{ref8} and DCPI-Depth (2025) \cite{ref7} on all seven metrics. These results demonstrate the improved generalization and support the effectiveness of selective cross-scale fusion and contextual modeling.

\subsection{Component ablation and qualitative analysis}

\label{ssec:experiment_4}

\noindent \hyperref[tab:component_ablation]{Table~\ref*{tab:component_ablation}} presents the ablation results for different combinations of the proposed components. Bi-CMamba and Uni-CMamba reduce SqRel from the baseline value of 0.632 to 0.624 and 0.621, respectively, while HAM reduces RMSE from 4.216 to 4.171. When all three components are combined, the model attains the best overall results. These results verify the functional complementarity of the two module families and the effectiveness of their joint modeling.

\hyperref[fig:experiments_4]{Fig.\ \ref*{fig:experiments_4}} presents a qualitative visual comparison on KITTI. The depth maps predicted by our method exhibit sharper object boundaries and more coherent local depth, which is consistent with the quantitative results in \hyperref[tab:quantitative_comparison]{Table~\ref*{tab:quantitative_comparison}}, and this visualization is intended to reflect the overall prediction quality.

\section{CONCLUSION}
\label{sec:majhead}

CMambaDepth combines bidirectional encoder-scale interaction, unidirectional decoder fusion, and hybrid attention. Quantitative comparisons on KITTI and DDAD, zero-shot cross-dataset generalization on NYUv2, and component ablations support the proposed design. Our model achieves 0.094 KITTI AbsRel and reduces NYUv2 AbsRel by 7.2\% relative to RA-Depth. In future work, we will explore multi-frame image sequences as input to strengthen generalization, and further reduce computational cost.

\clearpage
\clearpage
\begingroup
\fontsize{9}{10.5}\selectfont

\endgroup


\begin{thebibliography}{99}
\setlength{\itemsep}{0pt plus 1fill}\setlength{\parsep}{0pt}
\label{sec:refs}
\bibitem{ref1}
D. Eigen, C. Puhrsch, and R. Fergus, ``Depth map prediction from a single image using a multi-scale deep network,'' \textit{in Advances in Neural Information Processing Systems}, vol. 27, 2014.

\bibitem{ref2}
R. Garg, V. K. Bg, G. Carneiro, and I. Reid, ``Unsupervised CNN for single view depth estimation: Geometry to the rescue,'' in \textit{Computer Vision}\textit{ }\textit{-- }\textit{ECCV}\textit{ 201}\textit{6}, B. Leibe, J. Matas, N. Sebe, and M. Welling, Eds. Cham: Springer International Publishing, 2016, pp. 740--756.

\bibitem{ref3}
C. Godard, O. Mac Aodha, and G. J. Brostow, ``Unsupervised monocular depth estimation with left-right consistency,'' in \textit{2017 IEEE Conference on Computer Vision and Pattern Recognition (CVPR)}, Honolulu, HI, USA, 2017, pp. 6602--6611.

\bibitem{ref4}
T. Zhou, M. Brown, N. Snavely, and D. G. Lowe, ``Unsupervised learning of depth and ego-motion from video,'' in \textit{2017 IEEE Conference on Computer Vision and Pattern Recognition (CVPR)}, Honolulu, HI, USA, 2017, pp. 6612--6619.

\bibitem{ref5}
C. Godard, O. Mac Aodha, M. Firman, and G. J. Brostow, ``Digging into self-supervised monocular depth estimation,'' in \textit{2019 IEEE/CVF International Conference on Computer Vision (ICCV)}, Seoul, Korea (South), 2019, pp. 3827--3837.

\bibitem{ref6}
L. Sun, J.-W. Bian, H. Zhan, W. Yin, I. Reid, and C. Shen, ``SC-DepthV3: Robust self-supervised monocular depth estimation for dynamic scenes,''\textit{ IEEE Transactions on Pattern Analysis and Machine Intelligence}, vol. 46, no. 1, pp. 497--508, Jan. 2024.

\bibitem{ref7}
M. Zhang, Y. Feng, Q. Chen, and R. Fan, ``DCPI-Depth: Explicitly infusing dense correspondence prior to unsupervised monocular depth estimation,''\textit{ }\textit{IEEE Transactions on Image Processing}, vol. 34, pp. 4258--4272, 2025.

\bibitem{ref8}
C. Zhao et al., ``MonoViT: Self-supervised monocular depth estimation with a vision transformer,'' in \textit{2022 International Conference on 3D Vision (3DV)}, Prague, Czech Republic, 2022, pp. 668--678.

\bibitem{ref9}
M. He, L. Hui, Y. Bian, J. Ren, J. Xie, and J. Yang, ``RA-Depth: Resolution adaptive self-supervised monocular depth estimation,'' in \textit{Computer Vision}\textit{ }\textit{-- ECCV 20}\textit{22}, S. Avidan, G. Brostow, M. Cissé, G. M. Farinella, and T. Hassner, Eds. Cham: Springer Nature Switzerland, 2022, pp. 565--581.

\bibitem{ref11}
Y. Zhu, R. Ren, W. Dong, X. Li, and G. Shi, ``TSUDepth: Exploring temporal symmetry-based uncertainty for unsupervised monocular depth estimation,'' \textit{Neurocomputing}, vol. 600, Art. no. 128165, 2024.

\bibitem{ref14}
A. Gu and T. Dao, ``Mamba: Linear-time sequence modeling with selective state spaces,'' \textit{arXiv preprint}\textit{ arXi}\textit{v:2312.00752}, 2023.

\bibitem{ref15}
Y. Liu et al., ``VMamba: Visual state space model,'' \textit{in Advances in Neural Information Processing Systems}, vol. 37, 2024, pp. 103031--103063.

\pagebreak
\bibitem{ref16}
I. Grigore and C.-A. Popa, ``MambaDepth: Enhancing long-range dependency for self-supervised fine-structured monocular depth estimation,''\textit{ }\textit{arXiv preprint}\textit{ }\textit{arXiv}\textit{:}\textit{2406.04532}, 2024.

\bibitem{ref17}
M.-H. Guo, C.-Z. Lu, Z.-N. Liu, M.-M. Cheng, and S.-M. Hu, ``Visual attention network,'' \textit{Computational Visual Media}, vol. 9, no. 4, pp. 733--752, 2023.

\bibitem{ref18}
Q. Fan, H. Huang, M. Chen, H. Liu, and R. He, ``RMT: Retentive networks meet vision transformers,'' in \textit{2024 IEEE/CVF Conference on Computer Vision and Pattern Recognition (CVPR)}, Seattle, WA, USA, 2024, pp. 5641--5651.

\bibitem{ref19}
K. Sun, B. Xiao, D. Liu, and J. Wang, ``Deep high-resolution representation learning for human pose estimation,'' in \textit{2019 IEEE/CVF Conference on Computer Vision and Pattern Recognition (CVPR)}, Long Beach, CA, USA, 2019, pp. 5686--5696.

\bibitem{ref20}
K. He, X. Zhang, S. Ren, and J. Sun, ``Deep residual learning for image recognition,'' in \textit{2016 IEEE Conference on Computer Vision and Pattern Recognition (CVPR)}, Las Vegas, NV, USA, 2016, pp. 770--778.

\bibitem{ref23}
A. Geiger, P. Lenz, C. Stiller, and R. Urtasun, ``Vision meets robotics: The KITTI dataset,'' \textit{International Journal of Robotics Research}, vol. 32, no. 11, pp. 1231--1237, 2013.

\bibitem{ref24}
V. Guizilini, R. Ambrus, S. Pillai, A. Raventos, and A. Gaidon, ``3D packing for self-supervised monocular depth estimation,'' in \textit{2020 IEEE/CVF Conference on Computer Vision and Pattern Recognition (CVPR)}, Seattle, WA, USA, 2020, pp. 2482--2491.

\bibitem{ref25}
N. Silberman, D. Hoiem, P. Kohli, and R. Fergus, ``Indoor segmentation and support inference from RGBD images,'' in \textit{Computer Vision}\textit{ }\textit{-- }\textit{ECCV}\textit{ 201}\textit{2}, A. Fitzgibbon, S. Lazebnik, P. Perona, Y. Sato, and C. Schmid, Eds. Berlin, Heidelberg: Springer Berlin Heidelberg, 2012, pp. 746--760.

\bibitem{ref26}
D. P. Kingma and J. Ba, ``Adam: A method for stochastic optimization,'' \textit{arXiv preprint }\textit{arXiv:1412.6980}, 2014.

\bibitem{ref27}
J. Deng, W. Dong, R. Socher, L.-J. Li, K. Li, and L. Fei-Fei, ``ImageNet: A large-scale hierarchical image database,'' in \textit{2009 IEEE Conference on Computer Vision and Pattern Recognition}, Miami, FL, USA, 2009, pp. 248--255.

\bibitem{ref21}
D. Han, J. Shin, N. Kim, S. Hwang, and Y. Choi, ``TransDSSL: Transformer based depth estimation via self-supervised learning,'' in\textit{ IEEE Robotics and Automation Letters}, vol. 7, no. 4, pp. 10969--10976, Oct. 2022.

\bibitem{ref22}
X. Qin, Y. Zhu, L. Wang, X. Zhang, C. He, and Q. Dong, ``Self-supervised monocular depth learning from unknown cameras: Leveraging the power of raw data,'' \textit{Image and Vision Computing}, vol. 157, Art. no. 105505, 2025.

\pagebreak
\end{thebibliography}
\end{document}